\documentclass[runningheads]{llncs}
\usepackage{graphicx}
\usepackage{float}  
\usepackage{subcaption} 
\usepackage{tikz}
\usepackage{comment}
\usepackage{amsmath,amssymb} 
\usepackage{color}
\usepackage{caption}
\usepackage{subcaption}
\usepackage[accsupp]{axessibility}  

\usepackage[pagebackref,breaklinks,colorlinks]{hyperref}

\begin{document}
\pagestyle{headings}
\mainmatter
\def\ECCVSubNumber{4420}  

\title{PETR: Position Embedding Transformation for Multi-View 3D Object Detection} 

\titlerunning{Position Embedding Transformation for Multi-View 3D Object Detection}
%
\author{Yingfei Liu\thanks{Equal contribution.}\qquad Tiancai Wang$^{\star}$ \qquad Xiangyu Zhang \qquad Jian Sun
}
\authorrunning{Liu et al.}
%

\institute{MEGVII Technology \\
\email{\{liuyingfei,wangtiancai,zhangxiangyu,sunjian\}@megvii.com}}

\maketitle

\begin{abstract}
In this paper, we develop position embedding transformation (PETR) for multi-view 3D object detection. PETR encodes the position information of 3D coordinates into image features, producing the 3D position-aware features. Object query can perceive the 3D position-aware features and perform end-to-end object detection. PETR achieves state-of-the-art performance (\textbf{50.4\%} NDS and \textbf{44.1\%} mAP) on standard nuScenes dataset and ranks 1$st$ place on the benchmark. It can serve as a simple yet strong baseline for future research.
Code is available at \url{https://github.com/megvii-research/PETR}.
\keywords{Position embedding, transformer, 3D object detection}
\end{abstract}

\section{Introduction}
\begin{figure*}[t]
	\centering  
	\begin{subfigure}{.32\textwidth}
			\centering
			\includegraphics[width=\textwidth]{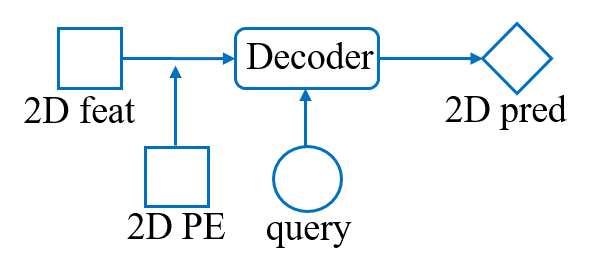}
			\caption{DETR}
		\end{subfigure}
	\begin{subfigure}{.32\textwidth}
			\centering
			\includegraphics[width=\textwidth]{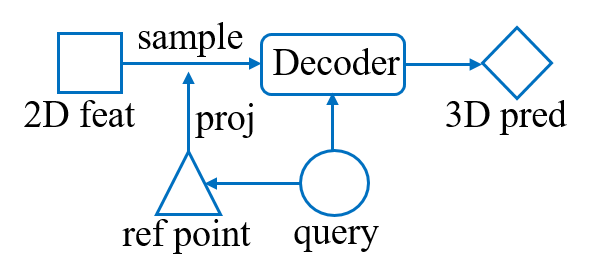}
			\caption{DETR3D}
		\end{subfigure}
	\begin{subfigure}{.32\textwidth}
			\centering
			\includegraphics[width=\textwidth]{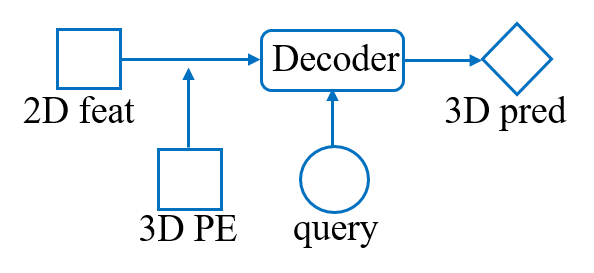}
			\caption{PETR}
		\end{subfigure}
	\caption{Comparison of DETR, DETR3D, and our proposed PETR. (a) In DETR, the object queries interact with 2D features to perform 2D detection. (b) DETR3D repeatedly projects the generated 3D reference points into image plane and samples the 2D features to interact with object queries in decoder. (c) PETR generates the 3D position-aware features by encoding the 3D position embedding (3D PE) into 2D image features. The object queries directly interact with 3D position-aware features and output 3D detection results.}  
	\label{arch_comparison}
\end{figure*}

3D object detection from multi-view images is appealing due to its low cost in autonomous driving system. Previous works~\cite{chen2016monocular,mousavian20173d,wang2021fcos3d,park2021dd3d,wang2022pgd} mainly solved this problem from the perspective of monocular object detection. Recently, DETR~\cite{carion2020detr} has gained remarkable attention due to its contribution on end-to-end object detection. In DETR~\cite{carion2020detr}, each object query represents an object and interacts with the 2D features in transformer decoder to produce the predictions (see Fig.~\ref{arch_comparison}(a)).
Simply extended from DETR~\cite{carion2020detr} framework, DETR3D~\cite{wang2022detr3d} provides an intuitive solution for end-to-end 3D object detection. The 3D reference point, predicted by object query, is projected back into the image spaces by the camera parameters and used to sample the 2D features from all camera views (see Fig.~\ref{arch_comparison}(b)). The decoder will take the sampled features and the queries as input and update the representations of object queries.

However, such 2D-to-3D transformation in DETR3D~\cite{wang2022detr3d} may introduce several problems. First, the predicted coordinates of reference point may not that accurate, making the sampled features out of the object region. Second, only the image feature at the projected point will be collected, which fails to perform the representation learning from global view. Also, the complex feature sampling procedure will hinder the detector from practical application.
Thus, building an end-to-end 3D object detection framework without the online 2D-to-3D transformation and feature sampling is still a remaining problem.

In this paper, we aim to develop a simple and elegant framework based on DETR~\cite{carion2020detr} for 3D object detection. We wonder if it is possible that we transform the 2D features from multi-view into 3D-aware features. In this way, the object query can be directly updated under the 3D environment. Our work is inspired by these advances in implicit neural representation~\cite{hu2019metasr,chen2021liff,mildenhall2020nerf}. 
In MetaSR~\cite{hu2019metasr} and LIFF~\cite{chen2021liff}, the high-resolution (HR) RGB values are generated from low-resolution (LR) input by encoding HR coordinates information into the LR features. In this paper, we try to transform the 2D features from multi-view images into the 3D representation by encoding 3D position embedding (see Fig.~\ref{arch_comparison}(c)).

To achieve this goal, the camera frustum space, shared by different views, is first discretized into meshgrid coordinates. The coordinates are then transformed by different camera parameters to obtain the coordinates of 3D world space. 
Then 2D image features extracted from backbone and 3D coordinates are input to a simple 3D position encoder to produce the 3D position-aware features.
The 3D position-aware features will interact with the object queries in transformer decoder and the updated object queries are further used to predict the object class and the 3D bounding boxes.  

The proposed PETR architecture brings many advantages compared to the DETR3D~\cite{wang2022detr3d}. It keeps the end-to-end spirit of original DETR~\cite{carion2020detr} while avoiding the complex 2D-to-3D projection and feature sampling. During inference time, the 3D position coordinates can be generated in an offline manner and served as an extra input position embedding. It is relatively easier for practical application.

To summarize, our contributions are:
\begin{itemize}
\item We propose a simple and elegant framework, termed PETR, for multi-view 3D object detection. The multi-view features are transformed into 3D domain by encoding the 3D coordinates. Object queries can be updated by interacting with the 3D position-aware features and generate 3D predictions.
\item A new 3D position-aware representation is introduced for multi-view 3D object detection. A simple implicit function is introduced to encode the 3D position information into 2D multi-view features.
\item Experiments show that PETR achieves state-of-the-art performance (\textbf{50.4\%} NDS and \textbf{44.1\%} mAP) on standard nuScenes dataset and ranks 1$st$ place on 3D object detection leaderboard. 
\end{itemize}
\section{Related Work}

\subsection{Transformer-based Object Detection}
Transformer~\cite{vaswani2017attention} is an attention block that widely applied to model the long-range dependency. In transformer, the features are usually added with position embedding, which provides the position information of the image~\cite{dosovitskiy2020image,wu2021rethinking,liu2021swin}, sequence~\cite{gehring2017convolutional,vaswani2017attention,devlin2018bert,dai2019transformer,yang2019xlnet}, and video~\cite{bertasius2021space,li2021improved,wu2022memvit}. Transformer-XL~\cite{dai2019transformer} uses the relative position embedding to encode the relative distance of the pairwise tokens. ViT~\cite{dosovitskiy2020image} adds the learned position embedding to the patch representations that encode distance of different patches. MViT~\cite{li2021improved} decomposes the distance computation of the relative position embedding and model the space-time structure.

Recently, DETR~\cite{carion2020detr} involves the transformer into 2D object detection task for end-to-end detection.
In DETR~\cite{carion2020detr}, each object is represented as an object query which interacts with 2D images features through transformer decoder. However, DETR~\cite{carion2020detr} suffers from the slow convergence. \cite{sun2021rethinking} attributes the slow convergence to the cross attention mechanism and designs a encoder-only DETR. Furthermore, many works accelerate the convergence by adding position priors. SMAC~\cite{gao2021fast} predicts 2D Gaussian-like weight map as spatial prior for each query.
Deformable DETR~\cite{zhu2020deformable} associates the object queries with 2D reference points and proposes deformable cross-attention to perform sparse interaction. \cite{wang2021anchor,meng2021conditional,liu2022dab} generate the object queries from anchor points or anchors that use position prior for fast convergence. Extended from DETR~\cite{zhu2020deformable}, SOLQ~\cite{dong2021solq} uses object queries to perform classification, box regression and instance segmentation simultaneously. 

\subsection{Vision-based 3D Object Detection}
Vision-based 3D object detection is to detect 3D bounding boxes from camera images. 
Many previous works~\cite{chen2016monocular,mousavian20173d,kehl2017ssd,ku2019monocular,simonelli2019disentangling,jorgensen2019monocular,brazil2019m3d,wang2021fcos3d,wang2022pgd} perform 3D object detection in the image view. 
M3D-RPN~\cite{brazil2019m3d} introduces the depth-aware convolution, which learns position-aware features for 3D object detection.
FCOS3D~\cite{wang2021fcos3d} transforms the 3D ground-truths to image view and extends FCOS~\cite{tian2019fcos} to predict 3D cuboid parameters. PGD~\cite{wang2022pgd} follows the FCOS3D~\cite{wang2021fcos3d} and uses a probabilistic representation to capture the uncertainty of depth. It greatly alleviates the depth estimation problem while introducing more computation budget and larger inference latency. DD3D~\cite{park2021dd3d} shows that depth pre-training on large-scale depth dataset can significantly improve the performance of 3D object detection. 


Recently, several works attempt to conduct the 3D object detection in 3D world space. 
OFT~\cite{roddick2018orthographic} and CaDDN~\cite{reading2021categorical} map the monocular image features into the bird’s eye view (BEV) and detect 3D objects in BEV space. ImVoxelNet~\cite{rukhovich2022imvoxelnet} builds a 3D volume in 3D world space and samples multi-view features to obtain the voxel representation. Then 3D convolutions and domain-specific heads are used to detect objects in both indoor and outdoor scenes. Similar to CaDDN~\cite{reading2021categorical}, BEVDet~\cite{huang2021bevdet} employs the Lift-Splat-Shoot~\cite{philion2020lift} to transform 2D multi-view features into BEV representation. With the BEV representation, a CenterPoint~\cite{yin2021center} head is used to detect 3D objects in an intuitive way.
Following DETR~\cite{carion2020detr}, DETR3D~\cite{wang2022detr3d} represents 3D objects as object queries. The 3D reference points, generated from object queries, are repeatedly projected back to all camera views and sample the 2D features.

BEV-based methods tend to introduce the Z-axis error, resulting in poor performance for other 3D-aware tasks (e.g., 3D lane detection). DETR-based methods can enjoy more benefits from end-to-end modeling with more training augmentations. Our  method is DETR-based that detects 3D objects in a simple and effective manner. We encode the 3D position information into 2D features, producing the 3D position-aware features. The object queries can directly interact with such 3D position-aware representation without projection error.

\subsection{Implicit Neural Representation}
Implicit neural representation (INR) usually maps the coordinates to visual signal by a multi-layer perceptron (MLP). It is a high efficiency way for modeling 3D objects~\cite{park2019deepsdf,chen2019learning,mescheder2019occupancy}, 3D scenes~\cite{mildenhall2020nerf,sitzmann2019scene,chabra2020deep,peng2020convolutional} and 2D images~\cite{hu2019metasr,chen2021liff,tancik2020fourier,sitzmann2020implicit}. NeRF~\cite{mildenhall2020nerf} employs a fully-connected network to denote a specific scene. To synthesize a novel view, the 5D coordinates along camera rays are input to the network as queries and output the volume density and view-dependent emitted radiance. In MetaSR~\cite{hu2019metasr} and LIFF~\cite{chen2021liff}, the HR coordinates are encoded into the LR features and HR images of arbitrary size can be generated. Our method can be regarded as an extension of INR in 3D object detection. The 2D images are encoded with 3D coordinates to obtain 3D position-aware features. The anchor points in 3D space are transformed to object queries by a MLP and further interact with 3D position-aware features to predict the corresponding 3D objects.


\section{Method}

\begin{figure*}[t]
	\centering  
	\includegraphics[height=5cm,width=12cm]{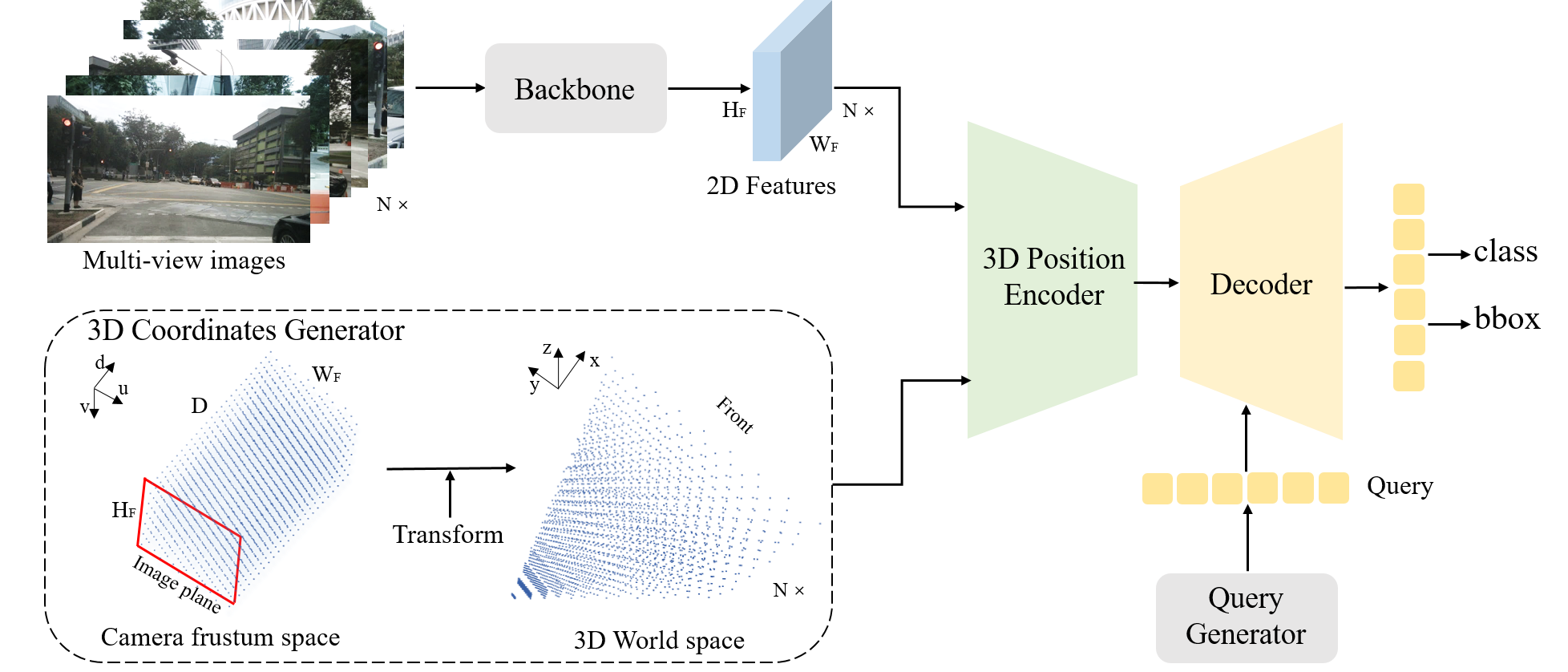}  
	\caption{The architecture of the proposed PETR paradigm. 
    The multi-view images are input to the backbone network (e.g. ResNet) to extract the multi-view 2D image features. In 3D coordinates generator, the camera frustum space shared by all views is discretized into a 3D meshgrid. The meshgrid coordinates are transformed by different camera parameters, resulting in the coordinates in 3D world space. Then 2D image features and 3D coordinates are injected to proposed 3D position encoder to generate the 3D position-aware features. Object queries, generated from query generator, are updated through the interaction with 3D position-aware features in transformer decoder. The updated queries are further used to predict the 3D bounding boxes and the object classes.
    }  
	\label{architecture}
\end{figure*}


\subsection{Overall Architecture}
Fig.~\ref{architecture} shows the overall architecture of the proposed PETR. Given the images $I=\{ I_i \in R^{3 \times H_I \times W_I}, i=1,2,\dots, N \}$ from $N$ views, the images are input to the backbone network (e.g. ResNet-50~\cite{he2016resnet}) to extract the 2D multi-view features  $F^{2d}=\{F^{2d}_i\in  R^{C \times H_F \times W_F}, i=1,2,\dots, N\}$. In 3D coordinates generator, the camera frustum space is first discretized into a 3D meshgrid. Then the coordinates of meshgrid are transformed by camera parameters and generate the coordinates in 3D world space. The 3D coordinates together with the 2D multi-view features are input to the 3D position encoder, producing the 3D position-aware features $F^{3d}=\{F^{3d}_i\in  R^{C \times H_F \times W_F}, i=1,2,\dots, N\}$. The 3D features are further input to the transformer decoder and interact with the object queries, generated from query generator. The updated object queries are used to predict the object class as well as the 3D bounding boxes.


\subsection{3D Coordinates Generator}
To build the relation between the 2D images and 3D space, we project the points in camera frustum space to 3D space since the points between these two spaces are one-to-one assignment. Similar to DGSN~\cite{chen2020dsgn}, we first discretize the camera frustum space to generate a meshgrid of size $(W_F,H_F,D )$. Each point in the meshgrid can be represented as $p^{m}_j = (u_j\times d_j, v_j\times d_j, d_j, 1)^T$, where $ (u_j, v_j) $ is a pixel coordinate in the image, $d_j$ is the depth value along the axis orthogonal to the image plane. Since the meshgrid is shared by different views, the corresponding 3D coordinate $p^{3d}_{i,j} = (x_{i,j}, y_{i,j}, z_{i,j}, 1)^T$ in 3D world space can be calculated by reversing 3D projection:
\begin{equation}\label{eq1}
p^{3d}_{i,j} = K^{-1}_{i} p^{m}_{j}
\end{equation}
where $K_i\in R^{4 \times 4}$ is the transformation matrix of $i$-th view that establish the transformation from 3D world space to camera frustum space. As illustrated in Fig. \ref{architecture}, the 3D coordinates of all views cover the panorama of the scene after the transformation. We further normalize the 3D coordinates as in Eq. \ref{eq2}.
\begin{equation}\label{eq2}
\left\{
\begin{aligned}
&x_{i,j} = &(x_{i,j}-x_{min}) / &(x_{max}-x_{min})\\
&y_{i,j} = &(y_{i,j}-y_{min}) / &(y_{max}-y_{min})\\
&z_{i,j} = &(z_{i,j}-z_{min}) / &(z_{max}-z_{min})
\end{aligned}
\right.
\end{equation}
where $[x_{min},y_{min},z_{min},x_{max},y_{max},z_{max}]$ is the region of interest (RoI) of 3D world space. The normalized coordinates of $H_F \times W_F \times D$ points are finally transposed as $P^{3d}=\{ P^{3d}_i \in R^{(D\times4) \times H_F\times W_F }, i=1,2,\dots, N \}$.

\begin{figure*}[t]
	\centering  
	\includegraphics[height=4cm,width=12cm]{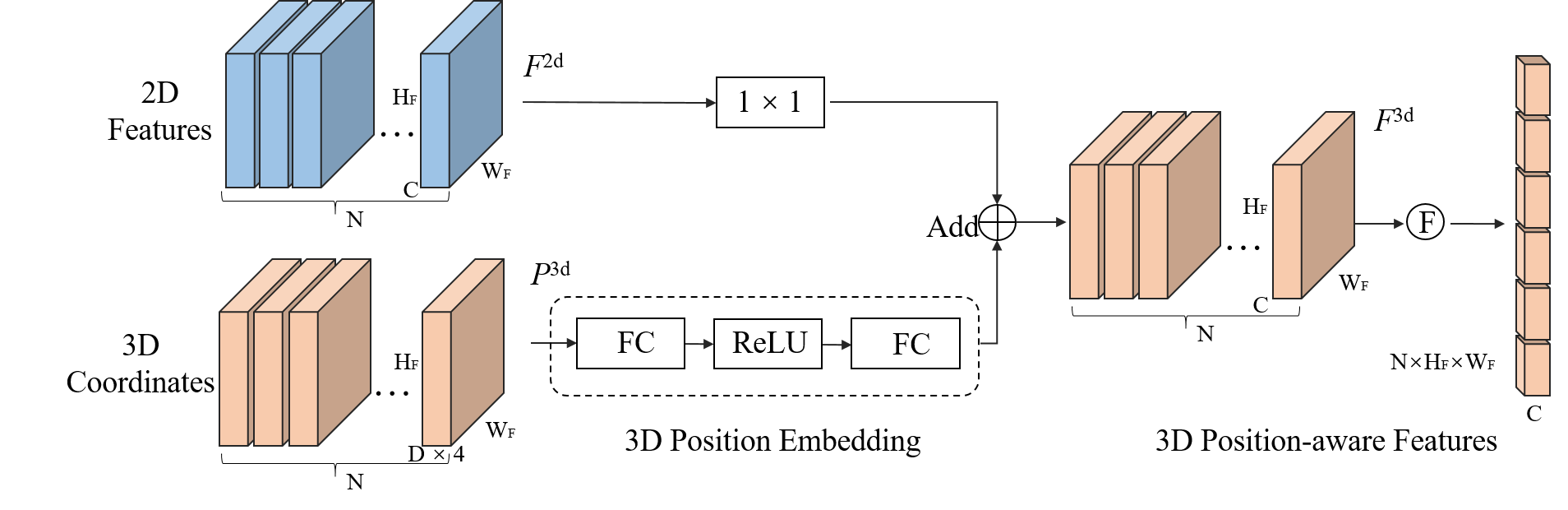}  
	\caption{Illustration of the proposed 3D Position Encoder. The multi-view 2D image features are input to a $1\times1$ convolution layer for dimension reduction. The 3D coordinates produced by the 3D coordinates generator are transformed into 3D position embedding by a multi-layer perception. The 3D position embeddings are added with the 2D image features of the same view, producing the 3D position-aware features. Finally, the 3D position-aware features are flattened and serve as the input of the transformer decoder.   $\textcircled{\scriptsize F}$ is the flatten operation.}  
	\label{fig2}
\end{figure*}

\subsection{3D Position Encoder}
The purpose of the 3D position encoder is to obtain 3D features $F^{3d}=\{F^{3d}_i\in  R^{C \times H_F \times W_F}, i=1,2,\dots, N\}$ by associating 2D image features $F^{2d}=\{F^{2d}_i\in \\R^{C \times H_F \times W_F}, i=1,2,\dots, N\}$ with 3D position information. Analogously to Meta SR~\cite{hu2019metasr}, the 3D position encoder can be formulated as:
\begin{equation}\label{eq3}
F^{3d}_i = \psi(F^{2d}_i,P^{3d}_i), \quad i=1,2,\dots, N
\end{equation}
where $\psi(.)$ is the position encoding function that is illustrated in Fig.~\ref{fig2}. Next, we describe the detailed implementation of $\psi(.)$.
Given the 2D features $F^{2d}$ and 3D coordinates $P^{3d}$, the $P^{3d}$ is first feed into a multi-layer perception (MLP) network and transformed to the 3D position embedding (PE). Then, the 2D features $F^{2d}$ is transformed by a $1\times1$ convolution layer and added with the 3D PE to formulate the 3D position-aware features. Finally, we flatten the 3D position-aware features as the key component of transformer decoder. 

\begin{figure*}[t]
	\centering  
	\includegraphics[height=3cm,width=12cm]{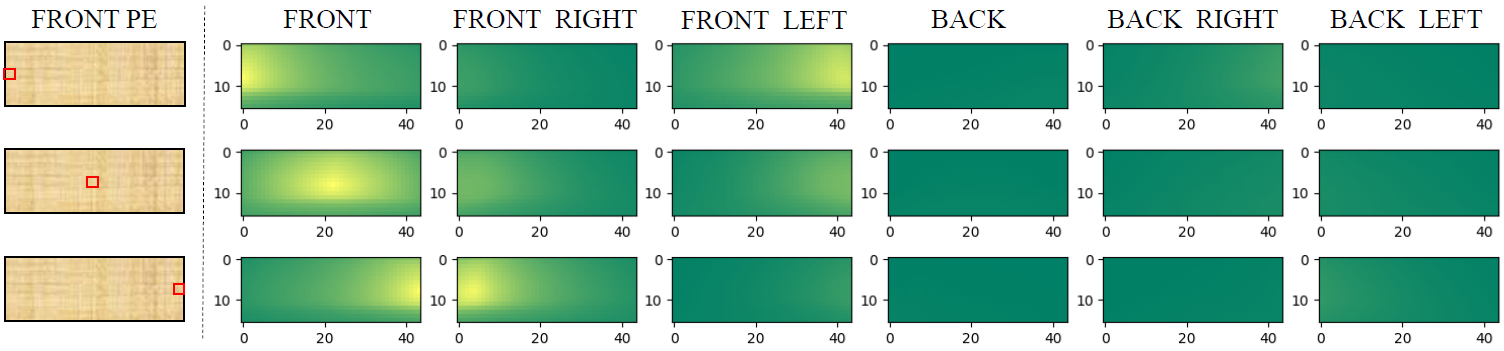}  
	\caption{3D position embedding similarity. The red points are selected positions in the front view. We calculated the similarity between the position embedding of these selected positions and all image views. It shows that the regions close to these selective points tend to have  higher similarity.}
	\label{similar}
\end{figure*}

\noindent \textbf{Analysis on 3D PE:} To demonstrate the effect of 3D PE, we randomly select the PE at three points in the front view and compute the PE similarity between these three points and all multi-view PEs. As shown in Fig. \ref{similar}, the regions close to these points tend to have the higher similarity. For example, when we select the left point in the front view, the right region of front-left view will have relatively higher response. It indicates that 3D PE implicitly establishes the position correlation of different views in 3D space.

\subsection{Query Generator and Decoder}
\subsubsection{Query Generator:}
Original DETR~\cite{carion2020detr} directly uses a set of learnable parameters as the initial object queries. Following Deformable-DETR~\cite{zhu2020deformable}, DETR3D~\cite{wang2022detr3d} predicts the reference points based on the initialized object queries. To ease the convergence difficulty in 3D scene, similar to Anchor-DETR~\cite{wang2021anchor}, we first initialize a set of learnable anchor points in 3D world space with uniform distribution from 0 to 1. Then the coordinates of 3D anchor points are input to a small MLP network with two linear layers and generate the initial object queries $Q_{0}$. In our practice, employing anchor points in 3D space can guarantee the convergence of PETR while adopting the setting in DETR or generating the anchor points in BEV space fail to achieve satisfying detection performance. For more details, please kindly refer to our experimental section.

\noindent \textbf{Decoder:}
For the decoder network, we follow the standard transformer decoder in DETR~\cite{carion2020detr}, which includes $L$ decoder layers. Here, we formulate the interaction process in decoder layer as:
\begin{equation}\label{eq4}
\begin{aligned}
Q_l = \Omega_l(F^{3d}, Q_{l-1}), \quad l=1, \dots, L
\end{aligned}
\end{equation}
where $\Omega_l$ is the $l$-th layer of the decoder. $Q_{l} \in R^{M\times C}$ is the updated object queries of $l$-th layer. $M$ and $C$ are the number of queries and channels, respectively. In each decoder layer, object queries interact with 3D position-aware features through the multi-head attention and feed-forward network. After iterative interaction, the updated object queries have the high-level representations and can be used to predict corresponding objects.

\subsection{Head and Loss} The detection head mainly includes two branches for classification and regression. The updated object queries from the decoder are input to the detection head and predict the probability of object classes as well as the 3D bounding boxes. 
Note that the regression branch predicts the relative offsets with respect to the coordinates of anchor points. For fair comparison with DETR3D, we also adopt the focal loss~\cite{lin2017focal} for classification and $L1$ loss for 3D bounding box regression. 
Let $y = (c,b)$ and $\hat{y} = (\hat{c},\hat{b})$ denote the set of ground truths and predictions, respectively. The Hungarian algorithm~\cite{kuhn1955hungarian} is used for label assignment between ground-truths and predictions. Suppose that $\sigma$ is the optimal assignment function, then the loss for 3D object detection can be summarized as:
\begin{equation}\label{eq5}
\begin{aligned}
L(y,\hat{y}) = \lambda_{cls} * L_{cls}(c,\sigma(\hat{c})) + L_{reg}(b,\sigma(\hat{b}))
\end{aligned}
\end{equation}

Here $L_{cls}$ denotes the focal loss for classification, $L_{reg}$ is $L1$ loss for regression. $\lambda_{cls}$ is a hyper-parameter to balance different losses.

\section{Experiments}
\subsection{Datasets and Metrics}
We validate our method on nuScenes benchmark~\cite{caesar2020nuscenes}. NuScenes is a large-scale multimodal dataset that is composed of data collected from 6 cameras, 1 lidar and 5 radars. The dataset has 1000 scenes and is officially divided into 700/150/150 scenes for training/validation/testing, respectively. Each scene has 20s video frames and is fully annotated with 3D bounding boxes every 0.5s. Consistent with official evaluation metrics, we report nuScenes Detection Score (NDS) and mean Average Precision (mAP), along with mean Average Translation Error (mATE), mean Average Scale Error (mASE), mean Average Orientation Error(mAOE), mean Average
Velocity Error(mAVE), mean Average Attribute Error(mAAE). 

\subsection{Implementation Details}
To extract the 2D features, ResNet~\cite{he2016resnet}, Swin-Transformer~\cite{liu2021swin} or VoVNetV2~\cite{lee2020centermask} are employed as the backbone network. The C5 feature (output of 5th stage) is upsampled and fused with C4 feature (output of 4th stage) to produce the P4 feature. The P4 feature with 1/16 input resolution is used as the 2D feature.
For 3D coordinates generation, we sample 64 points along the depth axis following the linear-increasing discretization (LID) in CaDDN~\cite{reading2021categorical}. We set the region to $[-61.2m, 61.2m]$ for the $X$ and $Y$ axis, and $[-10m, 10m]$ for $Z$ axis. The 3D coordinates in 3D world space are normalized to [0, 1]. Following DETR3D~\cite{wang2022detr3d}, we set $\lambda_{cls} = 2.0$ to balance classification and regression.

PETR is trained using AdamW~\cite{loshchilov2017decoupled} optimizer with weight decay of 0.01. The learning rate is initialized with $2.0\times10^{-4}$ and decayed with cosine annealing policy~\cite{loshchilov2016sgdr}. Multi-scale training strategy is adopted, where the shorter side is randomly chosen within [640, 900] and the longer side is less or equal to 1600. Following CenterPoint~\cite{yin2021center}, the ground-truths of instances are randomly rotated with a range of [$-22.5^{\circ}$, $22.5^{\circ}$] in 3D space.  All experiments are trained for 24 epochs (2x schedule) on 8 Tesla V100 GPUs with a batch size of 8. No test time augmentation methods are used during inference.

\setlength{\tabcolsep}{1pt}
\begin{table}[t!]
\begin{center}
\caption{Comparison of recent works on the nuScenes val set. The results of FCOS3D and PGD are fine-tuned and tested with test time augmentation. The DETR3D, BEVDet and PETR are trained with CBGS~\cite{zhu2019class}. $\dagger$ is initialized from an FCOS3D backbone.
}
\label{table:1}
\begin{tabular}{l|cc|ccccccc}
\hline\noalign{\smallskip}
Methods & Backbone & Size  & NDS$\uparrow$ & mAP$\uparrow$ & mATE$\downarrow$ & mASE$\downarrow$ & mAOE$\downarrow$ & mAVE$\downarrow$ & mAAE$\downarrow$ \\

\noalign{\smallskip}
\hline
\noalign{\smallskip}
 CenterNet&DLA  &  - &0.328 &0.306 &0.716 &0.264 &0.609 &1.426 &0.658   \\
 FCOS3D &Res-101  & 1600$\times$900  &0.415 &0.343 &0.725 &0.263 &0.422 &1.292 &0.153   \\
 PGD &Res-101 & 1600$\times$900  &0.428 &0.369  &0.683 &0.260 &0.439 &1.268 &0.185  \\
\hline
BEVDet &Res-50 & 1056$\times$384 &0.381 &0.304 &0.719 &0.272 &0.555 &0.903 &0.257 \\
BEVDet &Res-101 & 1056$\times$384 &0.389 &0.317 &0.704 &0.273 &0.531 &0.940 &0.250 \\
BEVDet &Swin-T & 1408$\times$512  &0.417 &0.349 &\textbf{0.637} &\textbf{0.269} &0.490 &0.914 &0.268 \\
PETR &Res-50 & 1056$\times$384 &0.381 &0.313 &0.768 &0.278 &0.564 & 0.923 &0.225  \\
PETR &Res-50 & 1408$\times$512 &0.403 &0.339 &0.748 &0.273 &0.539 &0.907 &0.203  \\
PETR &Res-101 & 1056$\times$384 &0.399 &0.333 &0.735 &0.275 &0.559 & 0.899 &0.205  \\
PETR &Res-101 & 1408$\times$512 &0.421 &0.357 &0.710 &0.270 &\textbf{0.490} &0.885 &0.224 \\
PETR &Swin-T & 1408$\times$512 &\textbf{0.431} &\textbf{0.361} &0.732 &0.273 &0.497 &\textbf{0.808} &\textbf{0.185}  \\
\hline
DETR3D$\dagger$ &Res-101  & 1600$\times$900  &0.434 &0.349 &0.716 &0.268 &\textbf{0.379} &0.842 &0.200 \\
PETR$\dagger$ &Res-101 & 1056$\times$384 &0.423 &0.347 &0.736 &0.269 &0.448 & 0.844 &0.202\\
PETR$\dagger$ &Res-101 & 1408$\times$512 &0.441 &0.366 &0.717 &0.267 &0.412 & \textbf{0.834} &\textbf{0.190} \\
PETR$\dagger$  &Res-101 & 1600$\times$900 &\textbf{0.442} &\textbf{0.370} &\textbf{0.711} &\textbf{0.267} &0.383 &  0.865 & 0.201 \\


\hline
\end{tabular}
\end{center}
\end{table}
\setlength{\tabcolsep}{1pt}

\subsection{State-of-the-art Comparison}
As shown in Tab. \ref{table:1}, we first compare the performance with state-of-the-art methods on nuScenes val set. It shows that PETR achieves the best performance on both NDS and mAP metrics. CenterNet~\cite{zhou2019objects}, FCOS3D~\cite{wang2021fcos3d} and PGD~\cite{wang2022pgd} are typical monocular 3D object detection methods. When compare with FCOS3D~\cite{wang2021fcos3d} and PGD~\cite{wang2022pgd}, PETR with ResNet-101~\cite{he2016resnet} surpasses them on NDS by 2.7\% and 1.4\%, respectively. However, PGD~\cite{wang2022pgd} achieves relatively lower mATE because of the explicit depth supervision. Besides, we also compare PETR with multi-view 3D object detection methods DETR3D~\cite{wang2022detr3d} and BEVDet~\cite{huang2021bevdet}, which detect 3D objects in a unified view. Since the DETR3D~\cite{wang2022detr3d} and BEVDet~\cite{huang2021bevdet} follow different settings on the image size and backbone initialization, we individually compare the PETR with them for fair comparison. Our method outperforms them 0.8\% and 1.4\% in NDS, respectively. 

\begin{figure*}[t]
	\centering  
	\begin{subfigure}{.49\textwidth}
			\centering
			\includegraphics[width=\textwidth]{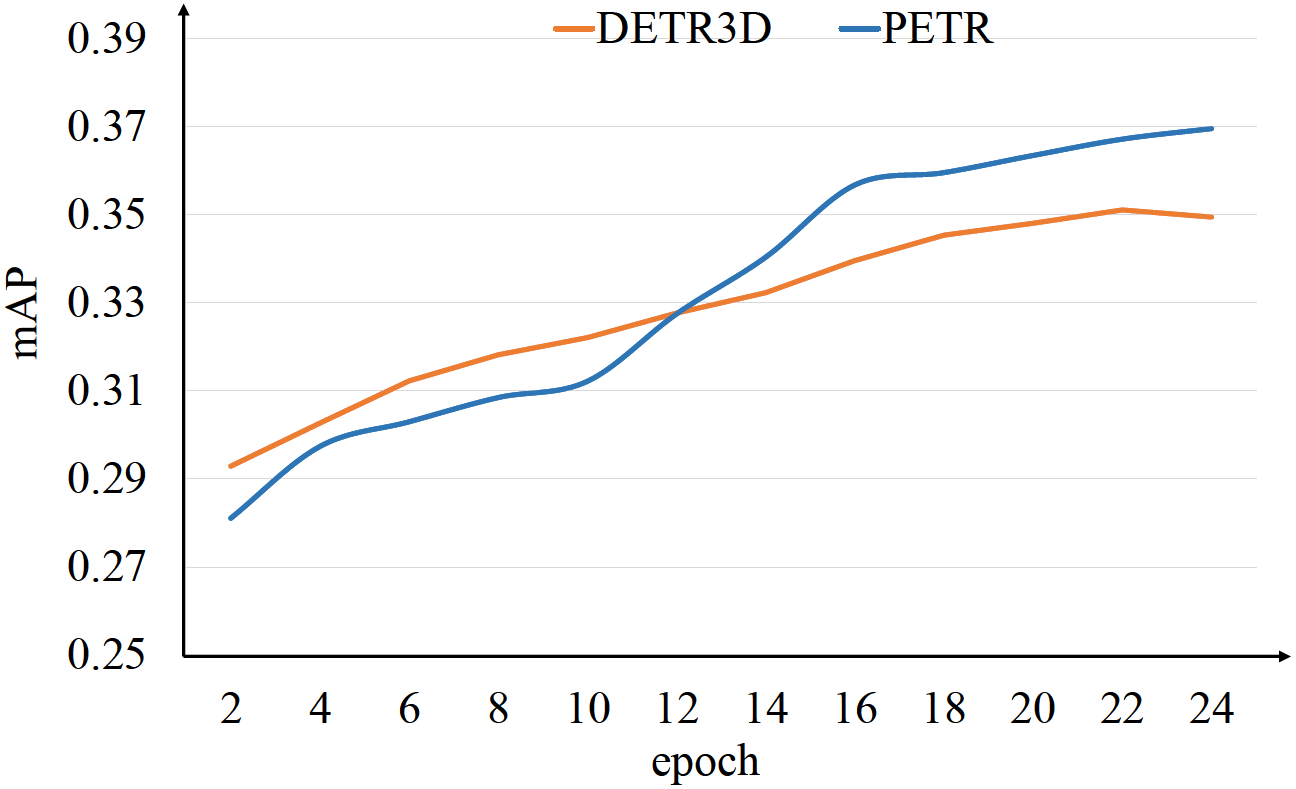}
			\caption{}
		\end{subfigure}
	\begin{subfigure}{.49\textwidth}
			\centering
			\includegraphics[width=\textwidth]{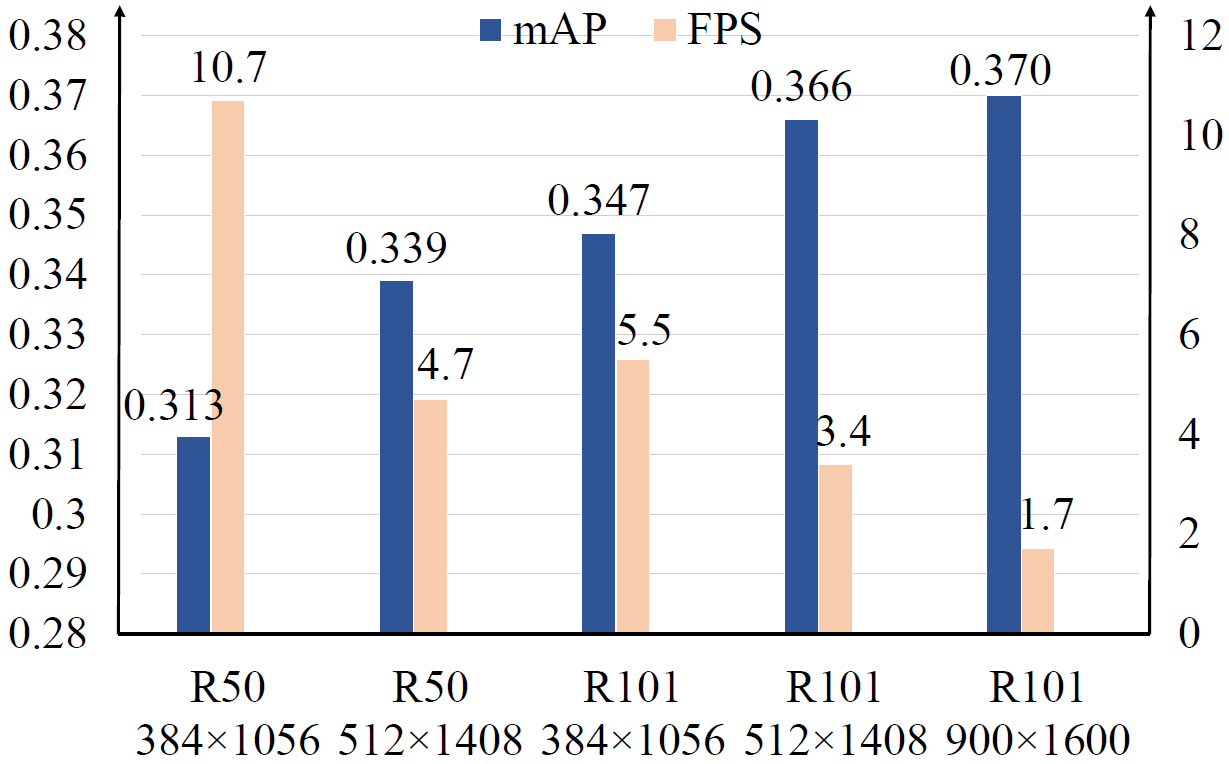}
			\caption{}
		\end{subfigure}
	\caption{Convergence and speed analysis on PETR. (a) The convergence comparison between PETR and DETR3D~\cite{wang2022detr3d}. PETR converges slower at initial stage and requires a relatively longer training schedule for fully convergence. (b) The performance and speed analysis with different backbones and input sizes. }
	\label{analysis}
\end{figure*}
\begin{table}[t!]
\begin{center}
\caption{Comparison of recent works on the nuScenes test set. $\ast$ are trained with external data.  $\ddagger$ is test time augmentation. }
\label{table:2}
\begin{tabular}{l|c|ccccccc}
\hline\noalign{\smallskip}
Methods & Backbone & NDS$\uparrow$ & mAP$\uparrow$ & mATE$\downarrow$ & mASE$\downarrow$ & mAOE$\downarrow$ & mAVE$\downarrow$ & mAAE$\downarrow$ \\

\noalign{\smallskip}
\hline
\noalign{\smallskip}
 FCOS3D$\ddagger$ & Res-101 &0.428 &0.358 &0.690 &0.249 &0.452 &1.434 &\textbf{0.124}  \\
 PGD$\ddagger$ & Res-101 &0.448 &0.386 &0.626 &0.245 &0.451 &1.509 &0.127  \\
 DD3D$\ast\ddagger$ & V2-99 &0.477 &0.418  &0.572 &0.249 &\textbf{0.368} &1.014 &\textbf{0.124}  \\
 DETR3D$\ast$ & V2-99 &0.479 &0.412  &0.641 &0.255 &0.394 &0.845 &0.133  \\
 BEVDet & Swin-S &0.463 &0.398 &0.556 &\textbf{0.239} &0.414 &1.010 &0.153  \\
 BEVDet$\ast$ & V2-99 &0.488 &0.424 &\textbf{0.524} &0.242 &0.373 &0.950 &0.148  \\
\hline
PETR & Res-101 &0.455 &0.391 &0.647 &0.251 &0.433 &0.933 &0.143 \\
PETR & Swin-T &0.450 &0.411 &0.664 &0.256 &0.522 &0.971 &0.137\\
PETR & Swin-S &0.481 &0.434 &0.641 &0.248 &0.437 &0.894 &0.143\\
PETR & Swin-B &0.483 &\textbf{0.445} &0.627 &0.249 &0.449 &0.927 &0.141\\					
PETR$\ast$ & V2-99 &\textbf{0.504} &0.441 &0.593 &0.249 &0.383 &\textbf{0.808} &0.132  \\
\hline
\end{tabular}
\end{center}
\end{table}
Tab.~\ref{table:2} shows the performance comparison on nuScenes test set. Our method also achieves the best performance on both NDS and mAP.
For fair comparison with BEVDet~\cite{huang2021bevdet}, PETR with Swin-S backbone is also trained with an image size of  $2112\times768$. It shows that PETR surpasses BEVDet~\cite{huang2021bevdet} by 3.6\% in mAP and 1.8\% in NDS, respectively. It is worth noting that PETR with Swin-B achieves a comparable performance compared to existing methods using external data. 
When using the external data, PETR with VOVNetV2~\cite{lee2020centermask} backbone achieves 50.4\% NDS and 44.1\% mAP. As far as we know, PETR is the first vision-based method that surpasses 50.0\% NDS.

We also perform the analysis on the convergence and detection speed of PETR. We first compare the convergence of DETR3D~\cite{wang2022detr3d} and PETR (see Fig.~\ref{analysis}(a)). PETR converges relatively slower than DETR3D~\cite{wang2022detr3d} within the first 12 epochs and finally achieves much better detection performance. It indicates that PETR requires a relatively longer training schedule for fully convergence. We guess the reason is that PETR learns the 3D correlation through global attention while DETR3D~\cite{wang2022detr3d} perceives the 3D scene within local regions. Fig.~\ref{analysis}(b) further reports the detection performance and speed of PETR with different input sizes. The FPS is measured on a single Tesla V100 GPU. For the same image size (e.g., 1056$\times$384), our PETR infers with 10.7 FPS compared to the BEVDet~\cite{huang2021bevdet} with 4.2 FPS. Note that the speed of BEVDet~\cite{huang2021bevdet} is measured on NVIDIA 3090 GPU, which is stronger than Tesla V100 GPU.
\begin{table}[t]
    \begin{center}
    \caption{The impact of 3D Position Embedding. 2D PE is the common position embedding used in DETR. MV is multi-view  position embedding to distinguish different views. 3D PE is the 3D position embedding proposed in our methods.}
    \label{table:3}
    \begin{tabular}{c|ccc|ccccccc}
        \hline\noalign{\smallskip}
        PE & 2D & MV & 3D& NDS$\uparrow$ & mAP$\uparrow$  & mATE$\downarrow$ & mASE$\downarrow$ & mAOE$\downarrow$ & mAVE$\downarrow$ & mAAE$\downarrow$\\
        \noalign{\smallskip}
        \hline
        \noalign{\smallskip}
        1&$\checkmark$& & &0.208 &0.069 &1.165 &0.290 &0.773 &0.936 &0.259\\
        2&$\checkmark$ &$\checkmark$& &0.224 &0.089 &1.165 &0.287 &0.738 &\textbf{0.929} &0.251 \\
        3& & &$\checkmark$&0.356 &0.305 &\textbf{0.835} &\textbf{0.238} &0.639 &0.971 &\textbf{0.237} \\
        4&$\checkmark$& &$\checkmark$&0.351 &0.305 &0.838 &0.283 &\textbf{0.633} &1.048 &0.256 \\
        5&$\checkmark$&$\checkmark$&$\checkmark$&\textbf{0.359} &\textbf{0.309} &0.844 &0.278 &0.653 &0.945 &0.241 \\
        \hline
        \end{tabular}
    \label{tab:array}
\end{center}
\end{table}

\setlength{\tabcolsep}{2.5pt}
\begin{table}[t!]
\begin{center}
\caption{Analysis of different methods to discrete the camera frustum space and different region of interest (ROI) ranges to normalized the 3D coordinates. UD is the Uniform discretization while LID is the linear-increasing discretization.}
\label{table:lid}
\begin{tabular}{c|c|c|c|ccc}
\hline\noalign{\smallskip}
Depth Range & $(x_{min},y_{min},z_{min},x_{max},y_{max},z_{max})$ & UD & LID  & NDS$\uparrow$ & mAP$\uparrow$ & mATE$\downarrow$\\
\noalign{\smallskip}
\hline
\noalign{\smallskip}
(1,51.2)&(-51.2, -51.2, -10.0, 51.2, 51.2, 10.0)&$\checkmark$ &  &0.352 &0.303 &0.862\\
(1,51.2)&(-51.2, -51.2, -5, 51.2, 51.2, 3) &$\checkmark$ &  &0.352 &0.305 &0.854\\
(1,61.2)&(-61.2, -61.2, -10.0, 61.2, 61.2, 10.0) &$\checkmark$ & &\textbf{0.358} &\textbf{0.308}  &\textbf{0.850}\\
(1,61.2)&(-61.2, -61.2, -5, 61.2, 61.2, 3) &$\checkmark$ &  &0.342 &0.297 &0.860\\
\hline
(1,51.2)&(-51.2, -51.2, -10.0, 51.2, 51.2, 10.0)& &$\checkmark$ &0.350 &\textbf{0.310} &0.843\\
(1,51.2)&(-51.2, -51.2, -5, 51.2, 51.2, 3) & &$\checkmark$ &0.355 &0.306 &\textbf{0.838}\\
(1,61.2)&(-61.2, -61.2, -10.0, 61.2, 61.2, 10.0) &&$\checkmark$ &\textbf{0.359} & 0.309 &0.839\\
(1,61.2)&(-61.2, -61.2, -5, 61.2, 61.2, 3) & &$\checkmark$ &0.346 &0.304 &0.842\\
\hline
\end{tabular}
\end{center}
\end{table}
\setlength{\tabcolsep}{1.4pt}

\begin{table}[t!]
    \begin{center}
    \caption{
    The ablation studies of different components in the proposed PETR.
    }
    \label{table:4}
    \setlength{\tabcolsep}{2.0pt}
    \begin{subtable}[t]{0.45\linewidth}
        \begin{tabular}{c|ccc}
        \hline\noalign{\smallskip}
        PE Networks& NDS$\uparrow$ & mAP$\uparrow$ & mATE$\downarrow$\\
        \noalign{\smallskip}
        \hline
        \noalign{\smallskip}
        None&0.311 &0.256 &1.00\\
        1$\times$1 ReLU 1$\times$1&\textbf{0.359} &\textbf{0.309} &\textbf{0.839}\\
        3$\times$3 ReLU 3$\times$3&0.017 &0.000 &1.054\\
        \hline
        \end{tabular}
        \caption{The network to generate the 3D PE. ``None'' means that the normalized 3D coordinates are directly used as 3D PE. }
    \end{subtable}
    \setlength{\tabcolsep}{2.5pt}
    \begin{subtable}[t]{0.45\linewidth}
        \begin{tabular}{c|ccc}
        \hline\noalign{\smallskip}
        Fusion Ways & NDS$\uparrow$ & mAP$\uparrow$ & mATE$\downarrow$\\
        \noalign{\smallskip}
        \hline
        \noalign{\smallskip}
        Add &\textbf{0.359} &\textbf{0.309} &0.839\\
        Concat&0.358 &0.309 &\textbf{0.832}\\
        Multiply&0.357 &0.303 &0.848\\
        \hline
        \end{tabular}
        \caption{Different ways to fuse the 2D multi-view features with 3D PE in the 3D position encoder.}
    \end{subtable}
    \setlength{\tabcolsep}{3.7pt}
    \begin{subtable}[t]{0.45\linewidth}
        \begin{tabular}{c|ccc}
        \hline\noalign{\smallskip}
        Anc-Points & NDS$\uparrow$ & mAP$\uparrow$ & mATE$\downarrow$\\
        \noalign{\smallskip}
        \hline
        \noalign{\smallskip}
        None &- &- & -\\
        Fix-BEV &0.337 &0.295 &0.852 \\
        Fix-3D &0.343 &0.303 &0.864\\
        Learned-3D&\textbf{0.359} &\textbf{0.309} &\textbf{0.839}\\
        \hline
        \end{tabular}
        \caption{
        ``None'' means no anchor points following DETR.
        ``Fix-BEV'' and ``Fix-3D'' mean the grid anchor points in BEV space and 3D space respectively. 
        }
    \end{subtable}
    \setlength{\tabcolsep}{3.4pt}
    \begin{subtable}[t]{0.45\linewidth}
        \begin{tabular}{c|ccc}
        \hline\noalign{\smallskip}
        Points-Num & NDS$\uparrow$ & mAP$\uparrow$ & mATE$\downarrow$\\
        \noalign{\smallskip}
        \hline
        \noalign{\smallskip}
        600&0.339 &0.300 &0.847\\
        900&0.351 &0.303 &0.860\\
        1200&0.354 &0.303 &0.854\\
        1500&\textbf{0.359} &\textbf{0.309} &\textbf{0.839}\\
        \hline
        \end{tabular}
        \caption{Results with different numbers of anchor points. We explored anchor point numbers ranging from 600 to 1500. More points perform better.}
    \end{subtable}
    \label{tab:array}
\end{center}
\end{table}

\subsection{Ablation Study}
In this section, we perform the ablation study on some important components of PETR. All the experiments are conducted using single-level C5 feature of ResNet-50 backbone without the CBGS~\cite{zhu2019class}.

\noindent \textbf{Impact of 3D Position Embedding.}
We evaluate the impact of different position embedding (PE) (see Tab. \ref{table:3}). When only the standard 2D PE in DETR is used, the model can only converge to 6.9\% mAP. Then we add the multi-view prior (convert the view numbers into PE) to distinguish different views and it brings a slight improvement. When only using the 3D PE generated by 3D coordinates, PETR can directly achieve 30.5\% mAP. It indicates that 3D PE provides a strong position prior to perceive the 3D scene. In addition, the performance can be improved when we combine the 3D PE with both 2D PE and multi-view prior. It should be noted that the main improvements are from the 3D PE and the 2D PE/multi-view prior can be selectively used in practice.

\noindent \textbf{3D Coordinate Generator.}
In 3D coordinates generator, the perspective view in camera frustum space is discretized into 3D meshgrid. The transformed coordinates in 3D world space are further normalized with a region of interest (RoI). Here, we explore the effectiveness of different discretization methods and RoI range (see Tab.~\ref{table:lid}). The Uniform discretization (UD) shows similar performance compared to the linear-increasing discretization (LID). We also tried several common ROI regions and the RoI range of $(-61.2m, -61.2m, -10.0m, 61.2m, 61.2m,$ $ 10.0m)$ achieves better performance than others.

\noindent \textbf{3D Position Encoder.}
The 3D position encoder is used to encode the 3D position into the 2D features. Here we first explore the effect of the multi-layer perception (MLP) that converts the 3D coordinates into 3D position embedding. It can be seen in Tab.~\ref{table:4}(a) that the network with a simple MLP can improve the performance by 4.8\% and 5.3\% on NDS and mAP compared to the baseline without MLP (aligning the channel number of 2D features to $D\times4$). When using two $3\times3$ convolution layers, the model will not converge as the $3\times3$ convolution destroys the correspondence between 2D feature and 3D position. Furthermore, we compare different ways to fuse the 2D image features with 3D PE in Tab.~\ref{table:4}(b). The concatenation operation achieves similar performance compared to addition while surpassing the multiply fusion.

\noindent \textbf{Query Generator.}
Tab.~\ref{table:4}(c) shows the effect of different anchor points to generate queries. Here, we compare four types of anchor points: ``None'', ``Fix-BEV'', ``Fix-3D'' and ``Learned-3D''. Original DETR (``None'') directly employs a set of learnable parameters as object queries without anchor points. The global feature of object query fail to make the model converge. ``Fix-BEV'' is the fixed anchor points are generated with the number of $39 \times 39$ in BEV space. ``Fix-3D'' means the fixed anchor points are with the number of $16 \times 16 \times 6$ in 3D world space. ``Learned-3D'' are the learnable anchor points defined in 3D space. We find the performance of both ``Fix-BEV'' and ``Fix-3D'' are lower than learned anchor points. We also explore the number of anchor points (see Tab.~\ref{table:4}(d)), which ranges from 600 to 1500. The model achieve the best performance with 1500 anchor points. Considering of the computation cost is increasing with the number of anchor points, we simply use 1500 anchor points to make a trade-off.

\begin{figure*}[t]
	\centering  
	\includegraphics[height=6.5cm,width=12cm]{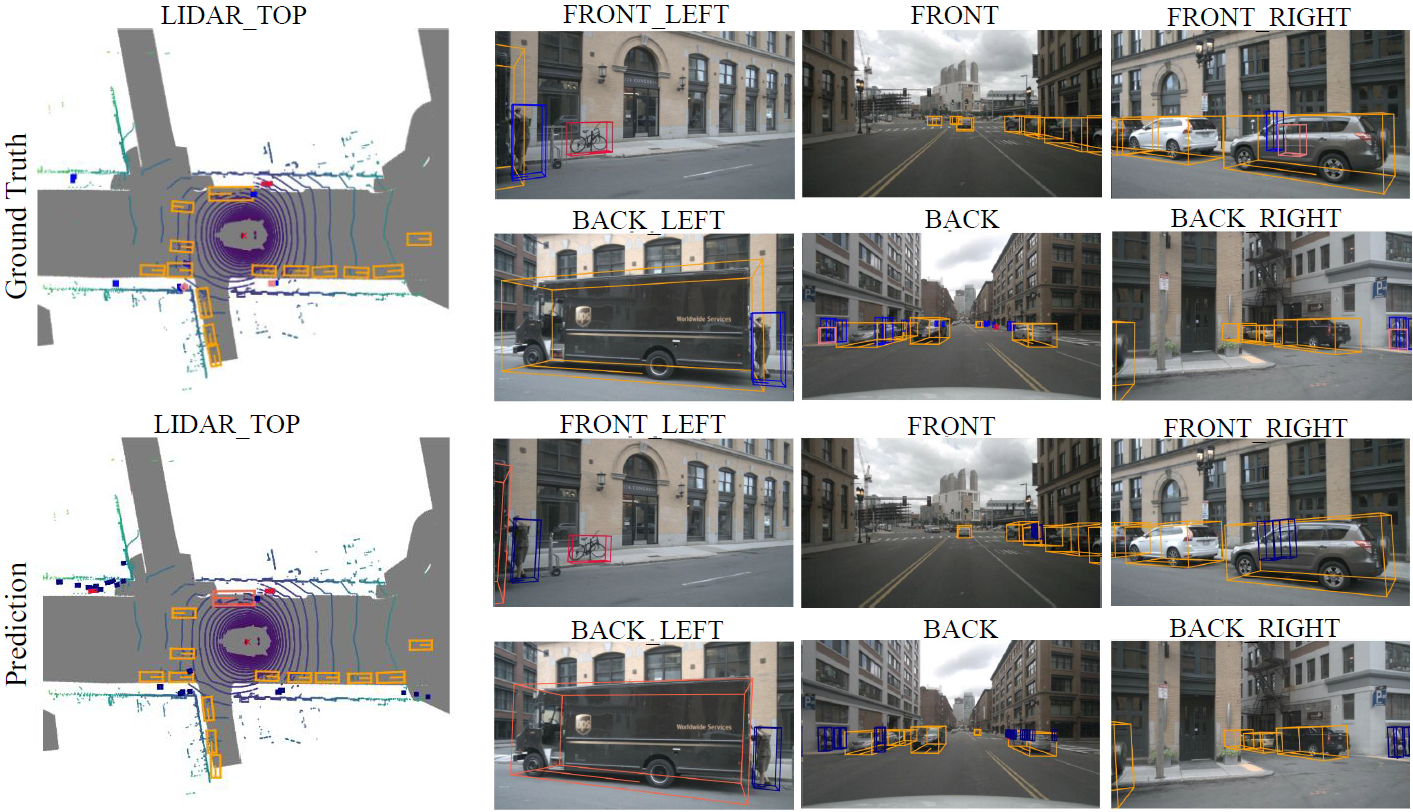}  
	\caption{Qualitative analysis of detection results in BEV and image views. The score threshold is 0.25, while the  backbone is  ResNet-101.
	The 3D bounding boxes are drawn with different colors to distinguish different classes. 
    }  
	\label{vis_result}
\end{figure*}
\begin{figure*}[t!]
	\centering  
	\includegraphics[height=3.5cm,width=12cm]{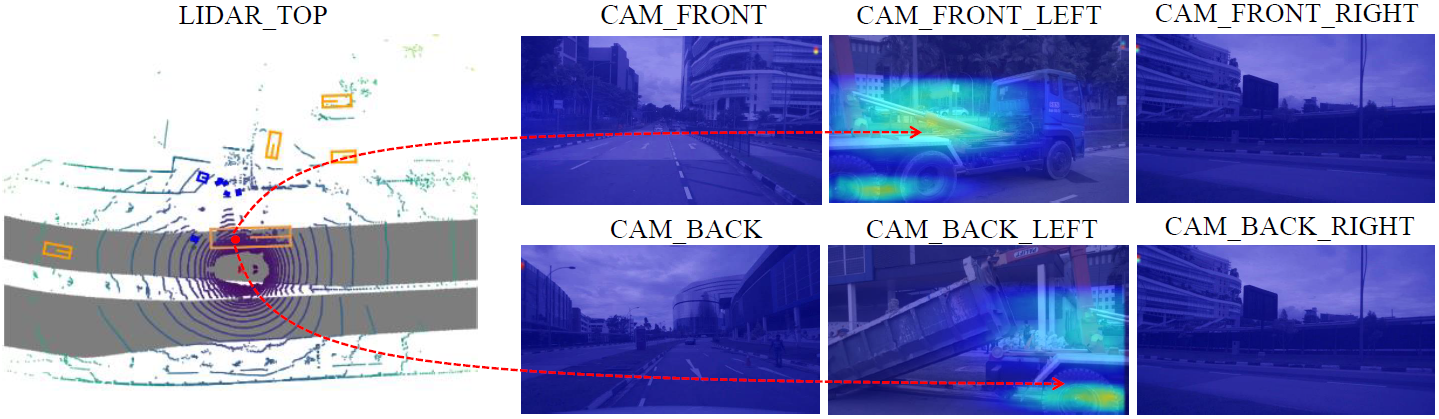}  
	\caption{Visualization of attention maps, generated from an object query (corresponding to the truck) on multi-view images. Both front-left and back-left views have a high response on the attention map.}
	\label{attention}
\end{figure*}
\begin{figure*}[t!]
	\centering  
	\includegraphics[height=5.75cm,width=12cm]{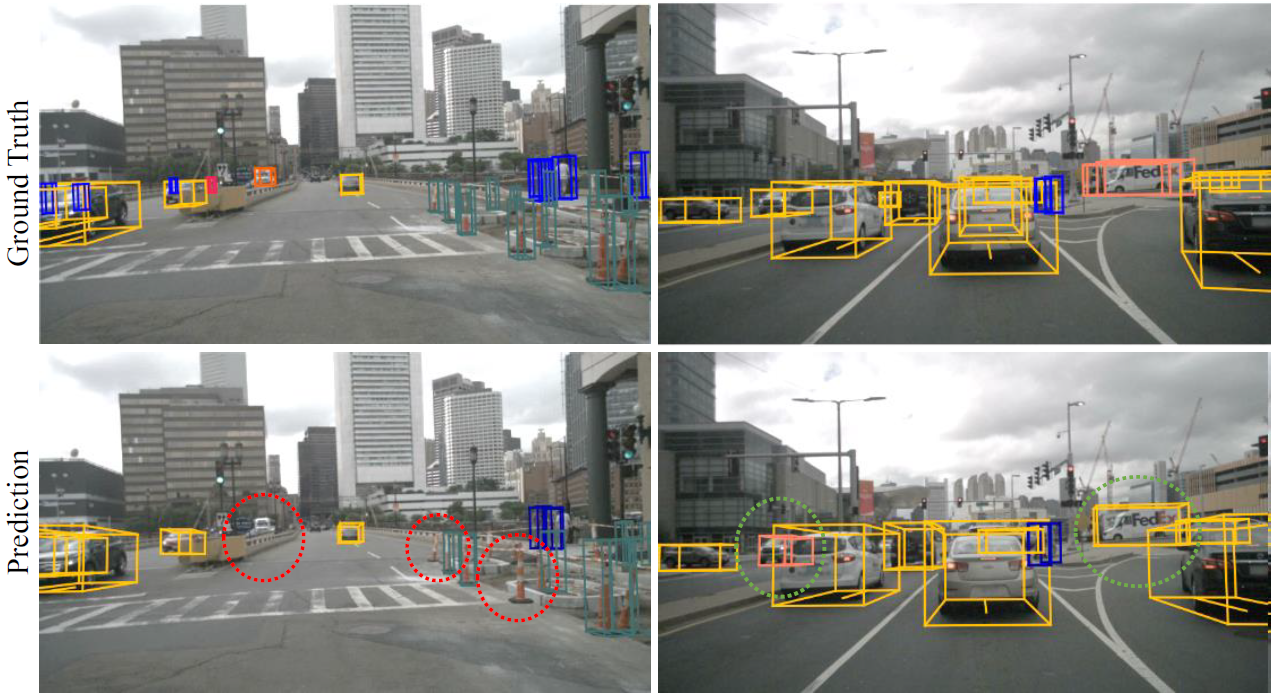}  
	\caption{Failure cases of PETR. We mark the failure cases by red and green circles. The red circles are some small objects that are not detected. The green circles are objects that are wrongly classified.}
	\label{badcase}
\end{figure*}
\subsection{Visualization}
Fig.~\ref{vis_result} shows some qualitative detection results. The 3D bounding boxes are projected and drawn in BEV space as well as image view. As shown in the BEV space, the predicted bounding boxes are close to the ground-truths. This indicates that our method achieves good detection performance. 
We also visualize the attention maps generated from an object query on multi-view images. As shown in Fig. \ref{attention}, the object query tends to pay attention to the same object, even in different views. It indicates that 3D position embedding can establish the position correlation between different views. Finally, we provide some failure cases (see Fig.~\ref{badcase}). The failure cases are marked by red and green circles. The red circles show some small objects that are not detected. 
The objects in green circle are wrongly classified. The wrong detection mainly occurs when different vehicles share high similarity on appearance.

\section{Conclusions}
The paper provides a simple and elegant solution for multi-view 3D object detection. By the 3D coordinates generation and position encoding, 2D features can be transformed into 3D position-aware feature representation. Such 3D representation can be directly incorporated into query-based DETR architecture and achieves end-to-end detection. It achieves state-of-the-art performance and can serve as a strong baseline for future research.

\subsubsection{Acknowledgements:} This research was supported by National Key R\&D Program of China (No. 2017YFA0700800) and Beijing Academy of Artificial Intelligence (BAAI).

%
%
\bibliographystyle{splncs04}
\bibliography{egbib}
\end{document}